%% file: main.tex
\input{_constants}

\cameraready 

\pdfoutput=1
\documentclass[10pt,twocolumn,letterpaper]{article}
\input{cvpr_header}

\unless\ifarxiv \myexternaldocument{_supplementary} \fi

\begin{document}
\title{\paperTitle}
\author{\authorBlock}
\maketitle





\input{00_abstract}
\input{01_intro}
\input{02_related}

\input{03_method}

\input{04_results}
\input{10_conclusion}

\section{Acknowledgment}
This work was supported in part by the Research Grants Council (RGC) of Hong Kong under the Research Impact Fund project R7003-21 and the Theme-based Research Scheme (TRS) Project T45-701-22-R. 

{\small
\bibliographystyle{ieeenat_fullname}
\bibliography{_main}
}

\ifarxiv \clearpage \appendix \input{12_appendix} \fi

\end{document}


\title{\paperTitle}
\author{\authorBlock}
\maketitlesupplementary

\appendix
\input{12_appendix}

{\small
\bibliographystyle{ieeenat_fullname}
\bibliography{11_references}
}

%% file: _constants.tex
\def\paperID{0000}
\def\confName{CVPR}
\def\confYear{2024}

\def\paperTitle{Co-designing a Sub-millisecond Latency Event-based Eye Tracking System with Submanifold Sparse CNN}

\def\authorBlock{
    Baoheng Zhang\thanks{Equal contribution} \qquad
    Yizhao Gao\footnotemark[1] \qquad
    Jingyuan Li \qquad
    Hayden Kwok-Hay So \qquad \\
    The University of Hong Kong \\
    {\tt\small \{bhzhang, yzgao, jyli, hso\}@eee.hku.hk}
}

\newif\ifreview 
\newif\ifarxiv 
\newif\ifcamera \newcommand{\cameraready}{\cameratrue}
\newif\ifrebuttal 

\newcommand{\figref}[1]{\figurename~{\ref{#1}}}
\newcommand{\tabref}[1]{Table~{\ref{#1}}}

\newcommand{\aname}{SEE\xspace}

%% file: cvpr_header.tex
\ifreview \usepackage[review]{cvpr} \fi
\ifarxiv \usepackage[pagenumbers]{cvpr} \fi
\ifrebuttal \usepackage[rebuttal]{cvpr} \fi
\ifcamera \usepackage{cvpr} \fi

\input{_macros}  

\usepackage{xr-hyper}

\makeatletter
\newcommand*{\addFileDependency}[1]{
  \typeout{(#1)}
  \@addtofilelist{#1}
  \IfFileExists{#1}{}{\typeout{No file #1.}}
}

\makeatother
\newcommand*{\myexternaldocument}[1]{
    \externaldocument{#1}
    \addFileDependency{#1.tex}
    \addFileDependency{#1.aux}
}

\definecolor{cvprblue}{rgb}{0.21,0.49,0.74}
\usepackage[pagebackref,breaklinks,colorlinks,citecolor=cvprblue]{hyperref}
\usepackage[capitalize]{cleveref}
\crefname{section}{Sec.}{Secs.}
\crefname{table}{Table}{Tables}
\crefname{figure}{Fig.}{Figs.}

\ifarxiv \crefname{appendix}{App.}{Apps.}
\else \crefname{appendix}{Suppl.}{Suppls.} \fi

%% file: _macros.tex
\usepackage{graphicx}	
\usepackage{amsmath}	
\usepackage{amssymb}	
\usepackage{booktabs}
\usepackage{times}
\usepackage{microtype}
\usepackage{epsfig}
\usepackage[table,xcdraw,dvipsnames]{xcolor}
\usepackage{caption}
\usepackage{float}
\usepackage{placeins}
\usepackage{color, colortbl}
\usepackage{stfloats}
\usepackage{enumitem}
\usepackage{tabularx}
\usepackage{xstring}
\usepackage{multirow}
\usepackage{xspace}
\usepackage{url}
\usepackage{subcaption}
\usepackage{xcolor}
\usepackage{nopageno}

\usepackage[hang,flushmargin]{footmisc}
\usepackage[accsupp]{axessibility}  
\ifcamera \usepackage[accsupp]{axessibility} \fi

% ...

\ifarxiv  \fi

\newcommand{\R}[1]{{%
    \textbf{%
        \ifstrequal{#1}{1}{\textcolor{red}{R#1}}{%
        \ifstrequal{#1}{2}{\textcolor{blue}{R#1}}{%
        \ifstrequal{#1}{3}{\textcolor{magenta}{R#1}}{%
        \ifstrequal{#1}{4}{\textcolor{teal}{R#1}}{%
                           \textcolor{cyan}{R#1}%
        }}}}%
    }%
}}

%% file: 00_abstract.tex
\begin{abstract}

Eye-tracking technology is integral to numerous consumer electronics applications, particularly in the realm of virtual and augmented reality (VR/AR). These applications demand solutions that excel in three crucial aspects: low-latency, low-power consumption, and precision. Yet, achieving optimal performance across all these fronts presents a formidable challenge, necessitating a balance between sophisticated algorithms and efficient backend hardware implementations. In this study, we tackle this challenge through a synergistic software/hardware co-design of the system with an event camera. Leveraging the inherent sparsity of event-based input data, we integrate a novel sparse FPGA dataflow accelerator customized for submanifold sparse convolution neural networks (SCNN). 
The SCNN implemented on the accelerator can efficiently extract the embedding feature vector from each representation of event slices by only processing the non-zero activations. Subsequently, these vectors undergo further processing by a gated recurrent unit (GRU) and a fully connected layer on the host CPU to generate the eye centers. 
Deployment and evaluation of our system reveal outstanding performance metrics. On the Event-based Eye-Tracking-AIS2024 dataset, our system achieves 81\% p5 accuracy, 99.5\% p10 accuracy, and 3.71 Mean Euclidean Distance with 0.7 ms latency while only consuming 2.29 mJ per inference. Notably, our solution opens up opportunities for future eye-tracking systems. Code is available at \url{https://github.com/CASR-HKU/ESDA/tree/eye_tracking}.


\end{abstract}

%% file: 01_intro.tex
\section{Introduction}
\label{sec:intro}

Eye tracking, the monitoring and analysis of eye movement and focus, provides valuable insights into visual attention, cognitive processes, and human-machine interaction. With applications ranging from psychology to marketing, eye tracking enables a deeper understanding of human behavior. For example, eye tracking plays a pivotal role in enhancing immersion and interaction in augmented and virtual reality (AR/VR)~\cite{VRrelated1, VRrelated2}. Also, it enables researchers to investigate visual perception, study information processing, optimize user interfaces, and enhance the design of human-computer interactions~\cite{eye_survey, ETsurvey2}.

A standard eye-tracking system is typically housed within embedded and wearable devices, necessitating comprehensive consideration of overall system performance, encompassing factors such as latency, power consumption, and precision.
Ensuring low latency guarantees real-time responsiveness and fluid interaction, thereby elevating the user experience. Low power consumption stands as a cornerstone for portable and wearable devices, facilitating prolonged usage periods without the need for frequent recharging. Precision remains paramount for capturing and analyzing eye movements accurately, enabling accurate gaze tracking for diverse applications.

However, achieving optimal performance in all three aspects poses a significant challenge. For instance, in past methodologies employing a frame-based camera in conjunction with a dense Deep Neural Networks (DNN) model, the system may incur noticeable latency and power consumption despite achieving satisfactory accuracy with the latency of 25 ms~\cite{eyeEfficient2}.


To this end, leveraging an event-based camera emerges as a promising solution. Unlike traditional cameras that capture images at fixed time intervals, event cameras detect light intensity changes on a per-pixel basis. This inherently sparse output significantly diminishes the data rate and potential backend processing demands. Moreover, the asynchronous nature of event cameras enables high temporal resolution, which offers a compelling pathway toward achieving low latency and precise eye movement tracking. Nevertheless, to fully exploit the sparsity and the high speed of event cameras, especially in deep learning models, exhibits great challenges. Off-the-shelf hardware platforms, such as GPU and CPU, usually fail to deliver satisfying performances in event processing.  

In this work, we provide an efficient \textbf{S}parse \textbf{E}vent-based \textbf{E}ye-tracking system called \textbf{\aname} by using hardware-software co-design that centralizes the idea of leveraging spatial sparsity. 
Our system adopts the submanifold sparse convolution neural network (SCNN) to efficiently extract feature vectors from the voxel grid representation of events. The SCNN model is deployed on an FPGA dataflow accelerator that can efficiently operate on sparse activations. The extracted features from the SCNN backbone are then fed into a combination of a GRU+FC module (implemented on the host CPU) to generate the normalized location of the eye center.

Our system is implemented end-to-end on an embedded FPGA SoC and extensively evaluated on the Event-based Eye-Tracking-AIS2024 dataset~\cite{3et, kaggle, wang2024ais_event}. Notably, our approach can achieve more than 98\% p10 accuracy with 0.7 ms - 0.94 ms inference latency. In comparison with an embedded GPU, our system achieves up to $15.4\times$ and $77.1\times$ speedup compared to standard and submanifold sparse convolution implementations, respectively.


%% file: 02_related.tex
\section{Related Work}
\label{sec:related}

\subsection{Eye tracking}
Eye tracking is a technology that involves monitoring and measuring the movement and focus of a person's eyes. It is used to understand and analyze human visual attention, gaze behavior, and eye movements. The process typically involves capturing and analyzing data related to the position, motion, and duration of eye fixations and saccades (rapid eye movements).

Traditional eye-tracking algorithms focus on using image processing methods to extract the center of the eye pupil. For example, \cite{traditioneye2} uses Harr-like features, K-means, and RANSAC-based ellipse fitting to recognize the pupil. ~\cite{traditioneye1} follows a three-step process, using contour segmentation to extract the pupil center. However, these methods are hard to deploy in real scenarios. As pointed out in \cite{larumbe2021accurate}, most of them are developed in controlled environments, and they always fail in some extreme environments, like changing view angles and illuminations.

With the development of deep learning, convolution neural networks (CNN) gradually become the most dominant method in solving computer vision tasks. CNN-based deep learning methods also become the mainstream to solve eye-tracking tasks, achieving much better performance than the traditional methods. For example, PupilNet~\cite{pupilnet} follows a coarse-to-fine mechanism, using a coarse CNN to obtain subregions and a fine CNN to generate the final response. ~\cite{choi2020eye} proposed a cascaded pipeline, using SSD~\cite{ssd} to detect the face, CycleGAN~\cite{cycleGAN} to remove the glasses, and FCN~\cite{FCN} to estimate the eye center location. 

Despite the notable advancements in accuracy, the efficiency of the algorithm remains a significant challenge. On one hand, the limited frame rate of traditional cameras hampers the system's capability to capture images at a high frequency. On the other hand, the proposed models demonstrate excessive complexity and high computational demands~\cite{alexNetEye, choi2020eye, xiang2022pupil, gou2019cascade}, making them difficult to deploy in a real-time system. While some other studies have focused on improving eye-tracking efficiency~\cite{efficient1, eyeEfficient2, lee2020deep}, the latency still exceeds 10 ms.
Event-based eye-tracking has recently gained attention as an emerging direction due to its advantages of low latency and low power consumption. However, despite the low latency offered by these approaches, recent works in the field primarily rely on traditional methods~\cite{eventEye1, eventEye2}, which often exhibit reduced accuracy.

\subsection{Acceleration of Event-based DNN Models}

As the era of deep learning, event vision has achieved remarkable progress in image classification~\cite{sun2022menet, peng2023get}, object detection~\cite{hamaguchi2023hierarchical, gehrig2023recurrent}, optical flow estimation~\cite{liu2023tma, ponghiran2023ofe}, etc. 
Despite the high-speed nature of the event sensor, many deep-learning-based event-based solutions are still suffering from the heavy computation. For instance, ~\cite{peng2023get} proposed using a Transformer model for image classification with over 10 ms latency, while some other Convolutional Neural Network (CNN) based solutions~\cite{hamaguchi2023hierarchical} can only achieve 7 ms inference using server-level GPUs like V100. 

Typically, the deep-learning-based solutions first convert the event streams into 2D images~\cite{image2d} or 3D voxels~\cite{voxel3d} and use dense models on GPU, losing the spatial sparsity of event streams. To address this challenge from a bottom-up approach, some previous works design special hardware accelerators to explore the sparsity~\cite{ESDA, nullhop} in the model. NullHop~\cite{nullhop} introduced an architecture that employs a binary bitmap to depict sparse activation on a per-layer basis, thereby avoiding unnecessary computations for zero values. ESDA~\cite{ESDA} proposes an all-on-chip sparse dataflow architecture on FPGA for low-latency and energy-efficient processing of event-based DNN models.

In this work, we build an efficient solution for eye-tracking problems by extending the ESDA framework with additional support of recurrent modules.  
Through the integration of software-hardware optimization techniques, our model achieves satisfactory accuracy and high hardware efficiency, with an overall latency less than 1 millisecond.

%% file: 03_method.tex
\section{Method}
\label{sec:method}

\subsection{Overall Architecture}

To address the eye-tracking problem efficiently, we propose SEE, a hardware-software co-optimization solution. On the software side, our model comprises a SCNN-based backbone for feature extraction, a GRU layer for temporal feature fusion, and a fully connected (FC) layer for eye center regression. Our hardware is heterogeneous, as the FPGA programmable fabric is used for SCNN acceleration and Arm Cortex-A series for GRU and FC layers. This heterogeneous architecture allows us to fully exploit the strengths of different hardware devices and deliver an overall low-latency performance. In addition,  we also employ hardware-software co-optimization to search for compact models with better tradeoffs between accuracy and hardware latency.

\subsection{Software design}

\subsubsection{Model Architecture}

\begin{figure*}
  \centering
  \includegraphics[width=\linewidth]{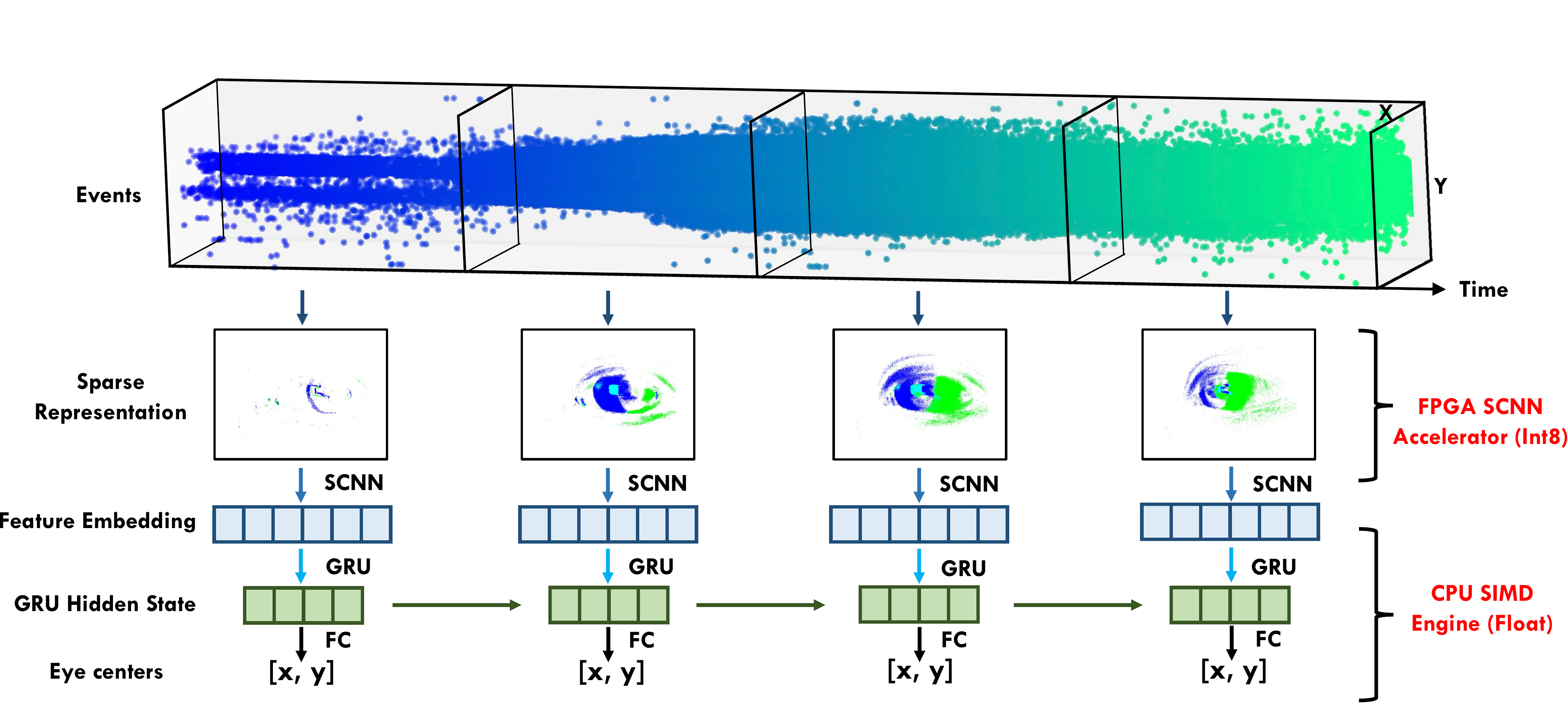}
  \caption{Software architecture: For an event stream, we partition it into multiple consecutive clips. These clips are then transformed into sparse voxel representations. Subsequently, an SCNN is used to generate feature embeddings, which will be then fed into a Gated Recurrent Unit (GRU) module. The GRU module generates the hidden state, and a Fully Connected (FC) layer regresses the eye centers.}
  \label{fig:sw_arch}
\end{figure*}

\aname follows the standard voxel grid representations as the input. As depicted in Figure \ref{fig:sw_arch}, the event clips in a fixed-time interval usually are spatially sparse, which means most of the pixels are zero. These sparse inputs are fed into the SCNN backbone to extract global features.
Subsequently, these features undergo further processing through a GRU layer, which captures the temporal information between event frames. The hidden features are then fed into the FC layer, yielding the normalized coordinates of the eye center location, ranging from 0 to 1. The actual eye location pixel coordinates can be obtained directly by multiplying these normalized coordinates with the height and width of the input size.


\subsubsection{Submanifold Sparse Convolution}
\label{sec:submanifold}
\begin{figure}[t]
  \centering
  \includegraphics[width=\columnwidth]{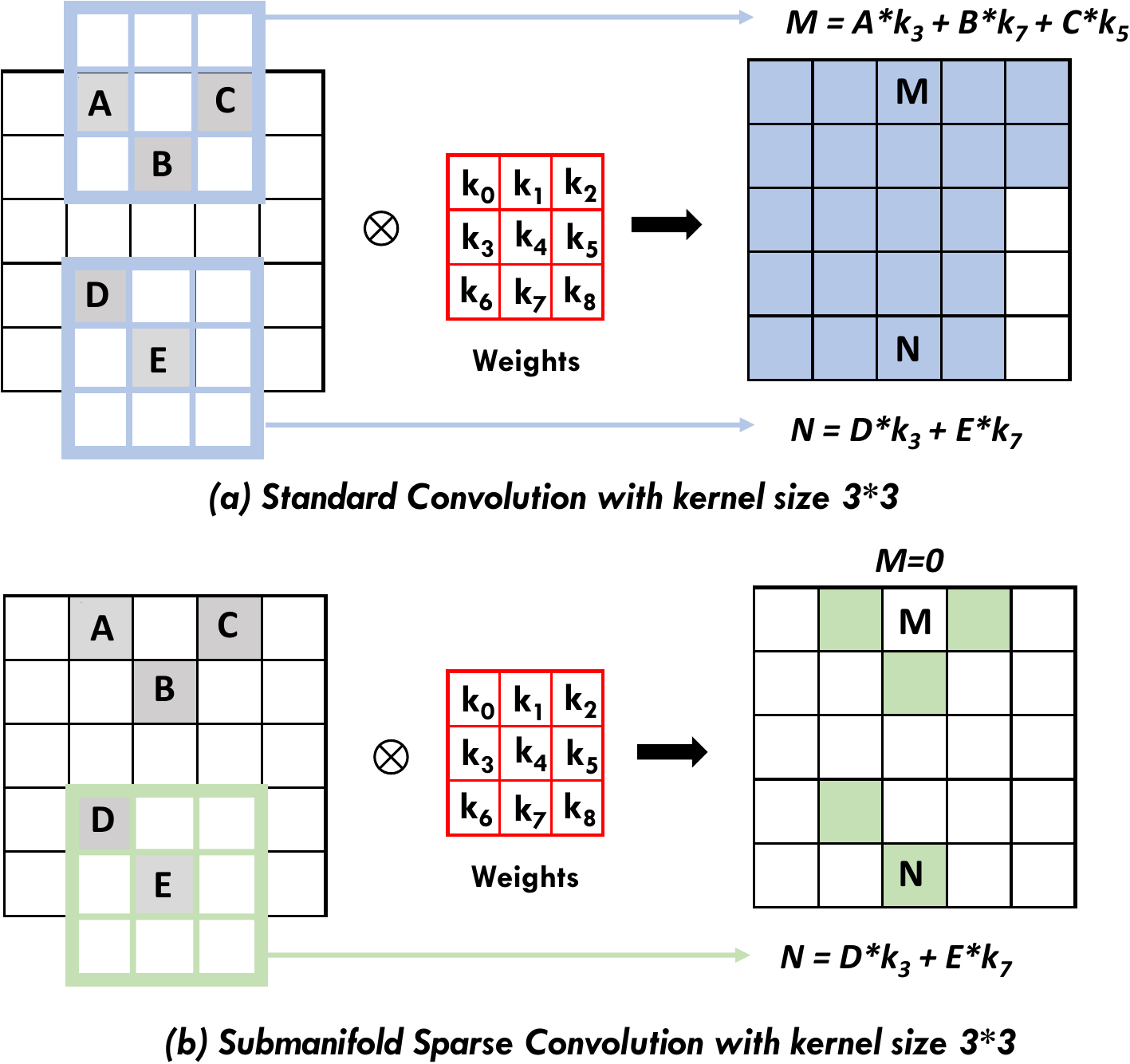}
  \caption{Standard and submanifold sparse convolution. For standard convolution, all the pixels in an image are processed by the kernels equally, leading to the dilation of spatial intensity. On the other hand, submanifold sparse convolution ensures that the output non-zero pixels locations to be identical as the input. 
  }
  \label{fig:submanifold}
\end{figure}

Convolutional layer (standard convolution) has been widely used in all kinds of deep learning architecture. However, the standard convolution algorithm suffers from a dilation effect when taking the spatially sparse input. Here, the spatial sparsity means some pixels in the input image or activations are completely zero for all the channels. As depicted in \figref{fig:submanifold}a, the dilation effect causes the output feature map to be much denser than the input. 

To address this issue and preserve spatial sparsity, we incorporate submanifold sparse convolution~\cite{mink} originally developed for point cloud networks. As illustrated in ~\figref{fig:submanifold}b, submanifold sparse convolution only outputs non-zeros on identical input non-zero locations. On a valid output location like N, the computation is exactly the same as that of standard convolution. 
In this way, the sparsity is preserved while meaningful information is propagated. 

By leveraging submanifold sparse convolution, we not only mitigate the dilation effect but also enhance the efficiency of inference by reducing unnecessary computations on irrelevant areas.

\subsubsection{Quantization}
\label{sec:quant}

Integer operations are more efficient and require fewer hardware resources for FPGA implementation. To deploy resource-efficient integer arithmetic on FPGA, we adopt HAWQv3~\cite{hawq} to fine-tune our model, which allows integer-only inference. Specifically, the models tuned by HAWQv3 only require integer multiplication, addition, and shift to be used in the whole computational graph.
In our experiments, we first train the model using the float32 data type as standard practice and perform fine-tuning by applying int8 quantization on both input $X$ and weights $W$ of the SCNN backbone. The quantization scheme can be expressed as:
\begin{equation}
    \begin{aligned}
    Y &= S_y\hat{Y} = W\times X = S_w\hat{W} \times S_x\hat{X} \\
    &\hat{Y} = \frac{S_wS_x}{S_y} (\hat{W} \times \hat{X}) = \frac{\hat{S}}{2^n}  (\hat{W} \times \hat{X})
    \end{aligned}
\end{equation}
where tensors with $\hat{}$ are in integer format. In this dyadic quantization scheme, the division of scaling factor $\frac{S_wS_x}{S_y}$ is replaced by an additional level of quantization with integer multiplication and shift operations (similar to a fixed point format). This allows simple hardware arithmetic to be deployed on the accelerator.  


\subsection{Hardware design}

\subsubsection{Overall Architecture}


The hardware diagram is shown in ~\figref{fig:hardware}, which is built upon a Xilinx Zynq UltraScale+ MPSoC device. The proposed hardware system primarily consists of two components: the sparse dataflow SCNN accelerator and the Arm Cortex-A53 processor host. The event-based input is initially fed into the SCNN accelerator to propagate through the submanifold sparse convolutional neural network backbone. Subsequently, the GRU and fully connected layers processes are executed by the host CPU with the Arm NEON SIMD (Single Instruction, Multiple Data) engine.

\subsubsection{FPGA SCNN Accelerator}
The FPGA SCNN accelerator adopts a dynamic sparse dataflow architecture introduced in \cite{ESDA}. This dataflow accelerator maps all the layers spatially on-chip and pipelines the computation of sparse activations for different modules. 
The dataflow modules share a unified token-feature streaming interface. A token $[.x,.y,.end]$ marks the current non-zero pixel coordinates. In a nutshell, the design of a dataflow module should comply with three principles: (1) it has the logic to resolve the next non-zero pixel coordinates; (2) it has the logic to compute the corresponding features at the next non-zero pixel; (3) the streaming order of non-zeros should follow the left-to-right, top-to-bottom manner. In this way, different model components, such as conv 1x1, conv 3x3, and pooling layers, can be implemented and cascaded in the dataflow manner, allowing sparse token-features to propagate throughout layers. 

\begin{figure*}
  \centering
  \includegraphics[width=\linewidth]{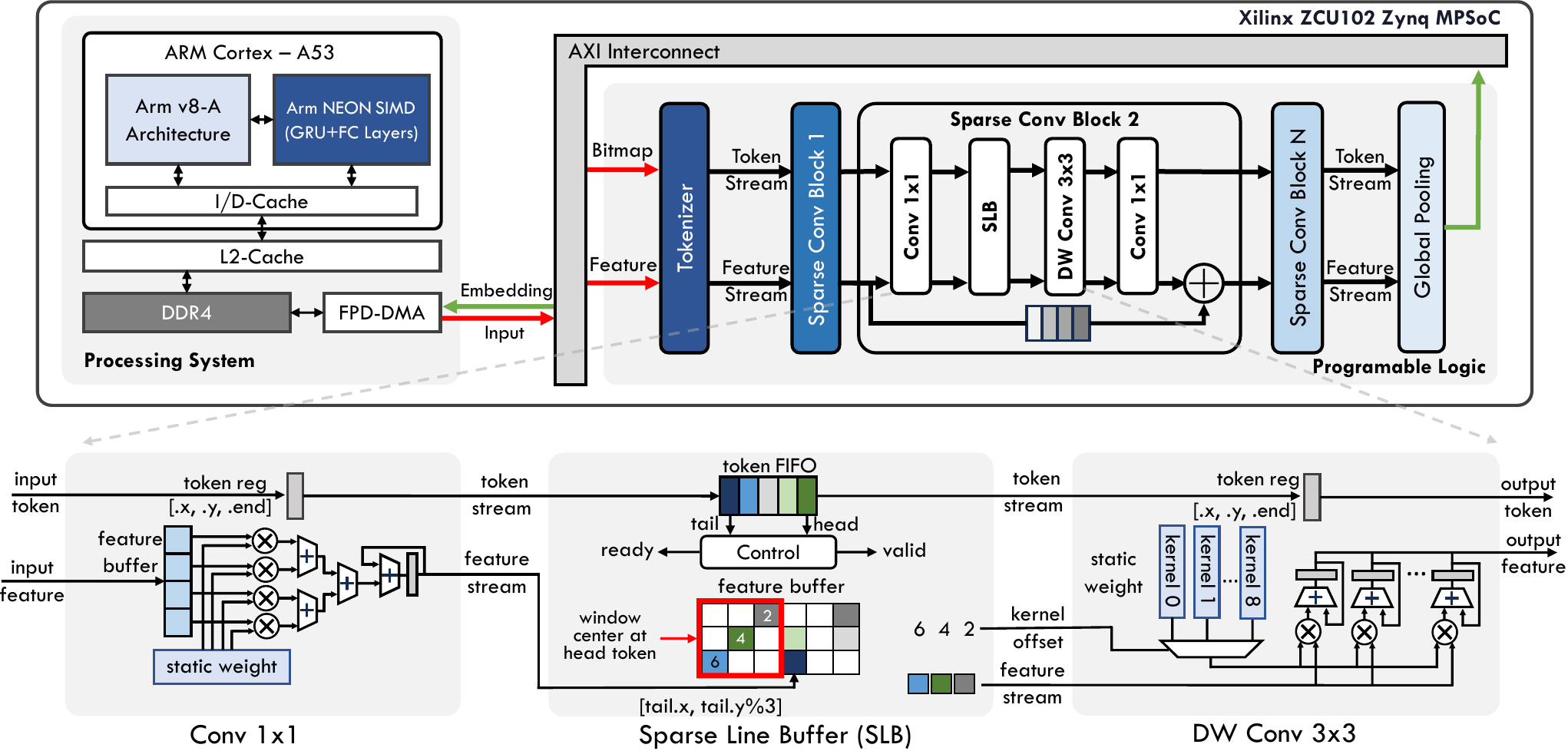}
  \caption{Heterogeneous Hardware architecture. The proposed hardware system primarily consists of the Arm Cortex-A53 acting as a processing system and the SCNN accelerator implemented on programable logic. The input to the SCNN accelerator is the sparse features and a binary bitmap to record the non-zero pixel locations. The GRU and fully connected layers are executed in the processing system with Arm NEON SIMD (Single Instruction, Multiple Data) engine.}
  \label{fig:hardware}
\end{figure*}

\figref{fig:hardware} shows an example diagram of a submanifold sparse conv 3x3 layer. It's composed of a Sparse Line Buffer (SLB) and a compute engine. Since the submanifold sparse convolution has identical input and output non-zero locations, the tokens can be simply buffered and reused by using a token FIFO. The head and tail tokens are used to control the read and write operations of the buffer. In addition, a kernel offset stream is used to exploit the sparsity within each 3x3 kernel. For example, only the offsets 2, 4, and 6 in the snapshot contain non-zero pixels. The kernel offsets subsequently serve as the index of the weight buffer in the later compute engine. 

This sparse dataflow scheme allows the non-zero information/features to be streamed and passed through different modules in the accelerator in an efficient pipeline. As discussed above, weight and activation are quantized into 8-bit integers to allow integer arithmetic units to be deployed while reducing memory consumption. Weights are stored in on-chip BRAM statically to reduce the off-chip communication in our design. However, one potential disadvantage of this approach is that model size can be limited by the on-chip buffer size. Fortunately, we have incorporated a co-optimization framework to trade off between model size and performance, which will be discussed in later sections.

\subsubsection{CPU SIMD Implementation of GRU+FC}
The main reason for deploying the GRU layer on the CPU is because its complex sigmoid activation functions are difficult to quantize and deploy on FPGA. Fortunately, the Arm SIMD engine has built-in floating point arithmetic units that are capable of handling the remaining GRU and FC layer within a reasonable time.  

The GRU and FC layers are implemented using the Eigen~\cite{eigenweb} C++ library. The computations involving vector operations are realized using several Arm NEON SIMD instructions. The compiled dynamic link library is packaged into Python and integrated with the host PYNQ (Python productivity for Zynq) platform~\cite{pynqPYNQ}.  

By combining the FPGA's parallel computing capability with the Arm Cortex-A series processor's efficient processing of SIMD operations, the proposed system optimizes the utilization of computational resources on Xilinx MPSoC platforms and maximizes both performance and efficiency for real-time eye-tracking applications.

\subsection{Software-hardware Co-optimization}

\begin{figure}
  \centering
  \includegraphics[width=\linewidth]{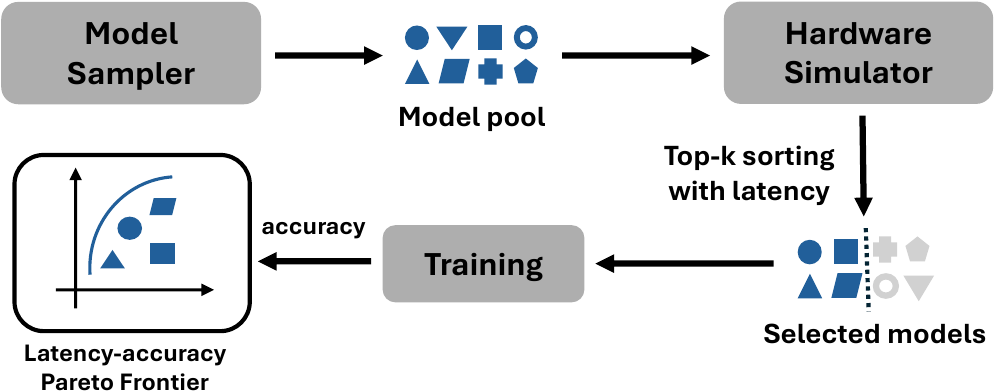}
  \caption{Network searching pipeline: We sample networks from a search pool and use a hardware simulator to select low-latency ones. After training these networks, we create a latency-accuracy Pareto frontier to show the trade-off between accuracy and latency.}
  \label{fig:search}
\end{figure}

Our system requires the entire backbone to fit the on-chip buffer. 
To achieve this objective, we have developed a software-hardware co-searching framework that aims to generate a compact network by considering both network complexity and hardware resource allocation, which is illustrated in  \figref{fig:search}. In this framework, we utilize MobileNetV2\cite{mobilenetv2} as a supernet and sample a large number of subnets. The searching space can be divided into four aspects: 1) the number of inverted bottleneck blocks, 2) the channel size of each block, 3) the ratio of expansion for each block, 4) the hidden feature size of the GRU layer. 

In the next stage, we select the candidate network architectures using a hardware simulator AGNA\cite{agna}. Given a model definition, the simulator uses a Geometric Programming method to estimate the latency based on the hardware constraints. 
Finally, the models with both lower estimated latency and feasible parameter sizes will be trained. In this pool of trained models, we select the ones lie within the pareto-frontier of accuracy and latency trade-offs.

%% file: 04_results.tex
\section{Experiments}
\label{sec:result}

\subsection{Implementation Details}

We conducted an efficiency verification of our design using a recently released dataset, the Event-based Eye-Tracking-AIS2024 dataset. The dataset comprises a total of 13 subjects, with each subject having 2-6 recording sessions. The subjects were instructed to perform activities belonging to 5 different classes, including random, saccades, read text, smooth pursuit, and blinks. We use the default split of the training and validation set.

The evaluation metrics are "Mean Euclidean Distance" (Dist.) and "p$k$ accuracy". Dist. is the average distance between ground truth and predicted locations, while "p$k$ accuracy" denotes the accuracy within a tolerance of $k$ pixels. Specifically, if the Euclidean distance between the ground truth and predicted locations is smaller than $k$ pixels, the sample is considered correct and vice versa. We utilize $k=5$ and $k=10$ to measure the prediction accuracy.


In terms of the hardware system, We implemented our hardware design with Vitis HLS 2020.2 and Vivado 2020.2. Then the proposed heterogeneous system is implemented and evaluated on a ZCU102 board with a Xilinx Zynq UltraScale+ MPSoC Device. 

In the subsequent sections, we refer to the models trained as the "\aname-series" models, denoting from \aname-A to \aname-D with different performance tradeoffs. MobileNetV2 (width multiplier = 0.5) is utilized as the baseline for comparison.

\subsection{Standard vs. Submanifold Sparse Convolution}


\begin{table}[]
\centering
\footnotesize
\caption{Accuracy between standard convolution and submanifold sparse convolution. }
\label{tab:Submanifold_acc}
\begin{tabular}{|c|cc|cc|}
\hline
      & \multicolumn{2}{c|}{p5 accuracy (\%)} & \multicolumn{2}{c|}{p10 accuracy (\%)} \\ \cline{2-5} 
            & \multicolumn{1}{c|}{Standard} & Sparse & \multicolumn{1}{c|}{Standard} & Sparse \\ \hline
MobileNetV2 & \multicolumn{1}{c|}{87.42}    & 87.63       & \multicolumn{1}{c|}{99.39}    & 99.46       \\ \hline
SEE-B & \multicolumn{1}{c|}{84.87}   & 85.21  & \multicolumn{1}{c|}{99.13}   & 98.86   \\ \hline
\end{tabular}
\end{table}

To demonstrate the model capability of submanifold sparse convolution, we carried out experiments to compare its performance with standard convolution. 
We use 2 different models including MobileNetV2 baseline and the \aname-B for the experiments, which are trained with standard and submanifold sparse convolution respectively. 
The results in \tabref{tab:Submanifold_acc} demonstrate that the p5 and p10 accuracy between the standard and submanifold implementations exhibit similarity. Specifically, the submanifold sparse convolution consistently achieves a comparable result in both p5 and p10 accuracy. However, this advantage comes with a notable increase in activation sparsity.




\subsection{Latency and Accuracy}
\label{subsec:latency_and_acc}
\begin{figure}[t]
  \centering
  \begin{subfigure}{\columnwidth}
    \centering
    \includegraphics[width=\columnwidth]{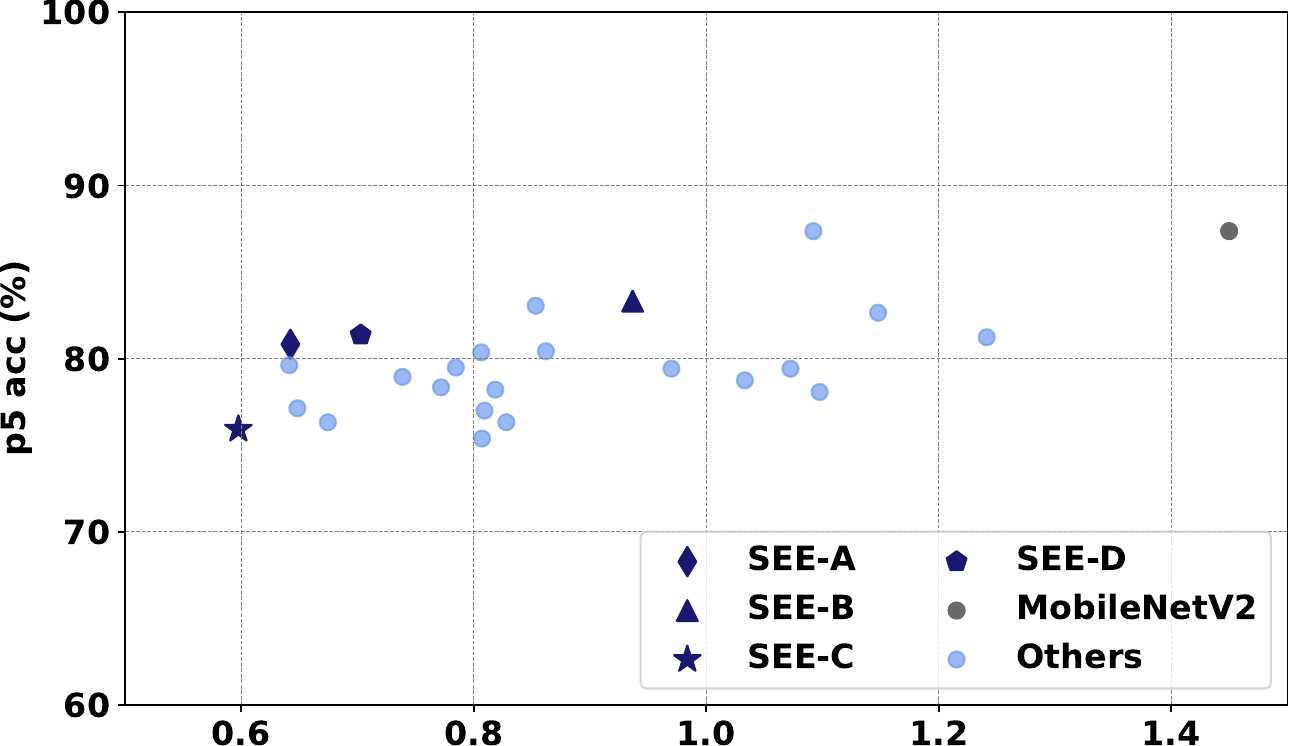}
    \caption{}
    \label{fig:subfig1}
  \end{subfigure}
  \begin{subfigure}{\columnwidth}
    \centering
    \includegraphics[width=\columnwidth]{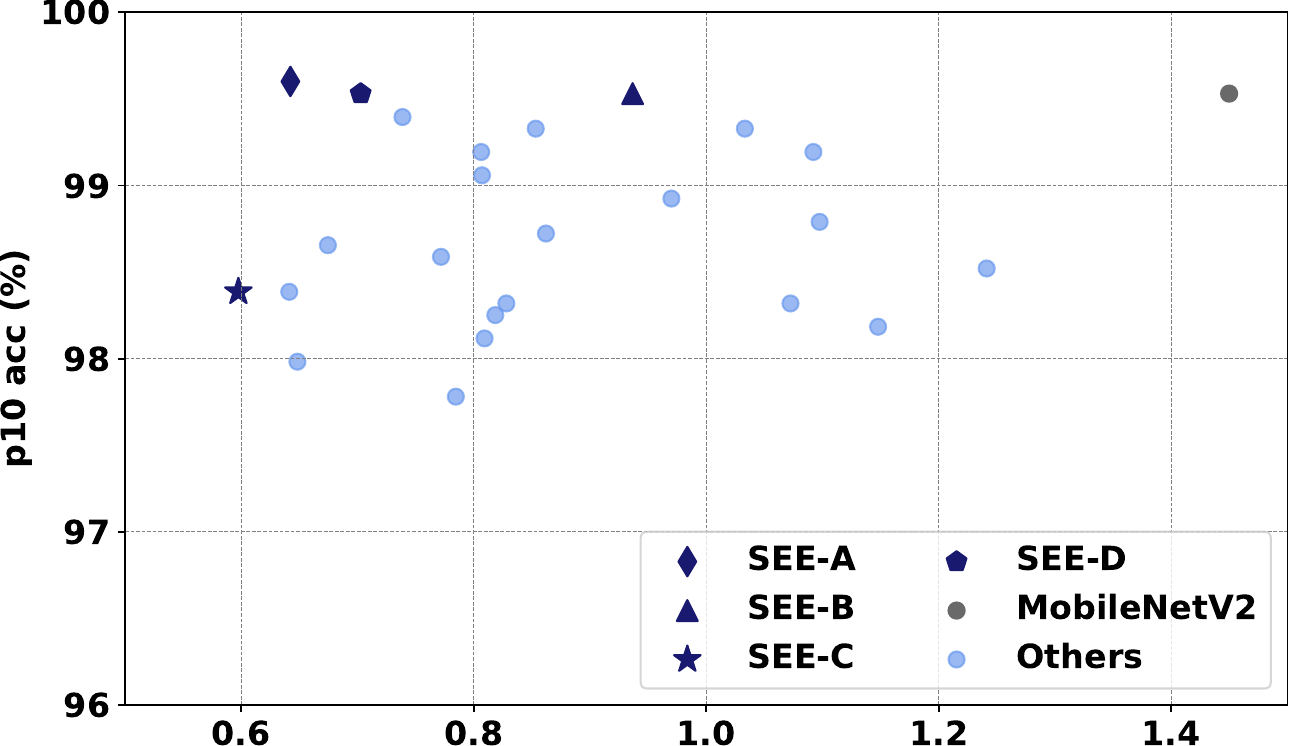}
    \caption{}
    \label{fig:subfig2}
  \end{subfigure}
  \begin{subfigure}{\columnwidth}
    \centering
    \includegraphics[width=\columnwidth]{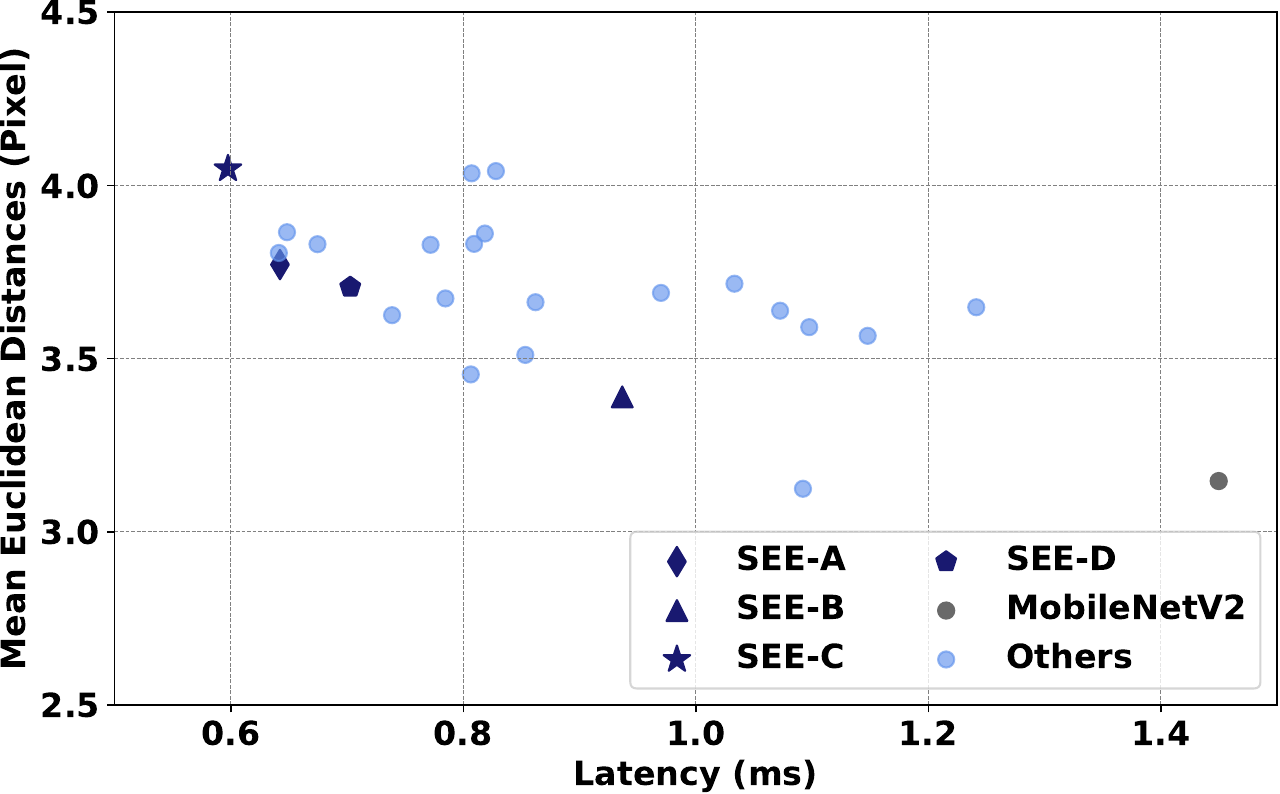}
    \caption{}
    \label{fig:subfig3}
  \end{subfigure}
  \caption{Accuracy vs Latency for sampled models. (a) p5 Accuracy vs. Latency. (b) p10 Accuracy vs. Latency. (c) Mean Euclidean Distances vs. Latency}
  \label{fig:latency_vs_acc}
\end{figure}

To demonstrate the effectiveness of our design, we follow our optimization flow in \figref{fig:search} to generate 20+ different models. The accuracy and latency results are plotted in \figref{fig:latency_vs_acc}, while the subgraph (a) and (b) show the latency with p5 and p10 accuracy respectively.

When evaluating with  p10 accuracy, we observe that the MobileNetV2 and the \aname-series networks achieve comparable high accuracies, mostly exceeding 98\%. While considering the p5 accuracy and the mean Euclidean distance, the baseline MobileNetV2 slightly outperforms the \aname-series models. This difference could be attributed to the higher number of network parameters since a larger model size generally provides more capacity to capture richer features.

In terms of efficiency, our selected \aname-series model significantly outperforms MobileNetV2 by a large margin. MobileNetV2 achieves a latency of 1.4 ms, which is more efficient than the previous work. However, our \aname-series model can even achieve a latency of less than 1 ms. Specifically, our \aname-D model achieves a comparable accuracy with MobileNetV2, with around $2\times$ speedup (0.7 ms vs. 1.45 ms). Our \aname-C model (0.6 ms) achieves around $2.5\times$ speedup over MobileNetV2 with only 1\% p10 accuracy drops. This highlights the capability of our \aname-framework to push more optimal latency accuracy trade-offs than baseline. 


\subsection{Hardware Implementation Details}

We also conduct further evaluations for our hardware implementation. We record the hardware-related parameters during the experiments, including resource utilization, power, and efficiency. The results are presented in Table \ref{tab:hardware_performance}. Notably, the \aname-series models consistently achieve low latency, with all inference times falling within the 1ms range. Additionally, our models consume lower power and demonstrate superior energy efficiency, as indicated by the reduced mJ per inference metric. These findings demonstrate the effectiveness of our approach in achieving low latency and low power consumption specifically for eye-tracking tasks. 

Our system provides a wide spectrum of performance tradeoffs. The SEE-A model obtains the highest p10 accuracy with more power, and the SEE-C model achieves the best overall latency and efficiency at the cost of slight degradation in accuracy. On the contrary, the SEE-B and the SEE-D models strike a more balanced tradeoff between accuracy and efficiency.

\begin{table*}[]
\centering
\footnotesize
\caption{Hardware Implementation Details.}
\begin{tabular}{cccccccccccccc}
\hline
 &
  \multicolumn{2}{c}{Accuracy (\%)} &
   &
   &
  \multicolumn{3}{c}{Latency (ms)} &
   &
   &
  \multicolumn{4}{c}{Utilization} \\ \cline{2-3} \cline{6-8} \cline{11-14} 
\multirow{-2}{*}{} &
  p5 &
  p10 &
  \multirow{-2}{*}{\begin{tabular}[c]{@{}c@{}}Dist.\\ (Pixel)\end{tabular}} &
  \multirow{-2}{*}{\begin{tabular}[c]{@{}c@{}}\#\\ Param.\end{tabular}} &
  SCNN &
  GRU\&FC &
  Total &
  \multirow{-2}{*}{\begin{tabular}[c]{@{}c@{}}Power\\ (W)\end{tabular}} &
  \multirow{-2}{*}{\begin{tabular}[c]{@{}c@{}}Efficiency\\ (mJ/inf.)\end{tabular}} &
  DSP &
  BRAM &
  FF &
  LUT \\ \hline
MobileNetV2 &
  {\color[HTML]{000000} \textbf{87.36}} &
  {\color[HTML]{000000} 99.53} &
  {\color[HTML]{000000} \textbf{3.15}} &
  797K &
  0.73 &
  0.72 &
  1.45 &
  4.36 &
  3.23 &
  2123 &
  1685 &
  213K &
  214K \\ \hline
SEE-A &
  {\color[HTML]{000000} 80.83} &
  {\color[HTML]{000000} \textbf{99.60}} &
  {\color[HTML]{000000} 3.77} &
  465K &
  \textbf{0.49} &
  0.15 &
  {\color[HTML]{000000} 0.64} &
  4.05 &
  1.99 &
  2003 &
  1287 &
  114K &
  166K \\
SEE-B &
  {\color[HTML]{000000} 83.32} &
  {\color[HTML]{000000} 99.53} &
  {\color[HTML]{000000} 3.39} &
  372K &
  0.79 &
  0.15 &
  {\color[HTML]{000000} 0.94} &
  4.17 &
  3.28 &
  2067 &
  1547 &
  117K &
  170K \\
SEE-C &
  \cellcolor[HTML]{FFFFFF}{\color[HTML]{000000} 75.92} &
  \cellcolor[HTML]{FFFFFF}{\color[HTML]{000000} 98.39} &
  \cellcolor[HTML]{FFFFFF}{\color[HTML]{000000} 4.05} &
  180K &
  \textbf{0.49} &
  \textbf{0.11} &
  {\color[HTML]{000000} \textbf{0.60}} &
  \textbf{3.86} &
  \textbf{1.88} &
  1880 &
  1001 &
  94K &
  135K \\
SEE-D &
  {\color[HTML]{000000} 81.37} &
  {\color[HTML]{000000} 99.53} &
  {\color[HTML]{000000} 3.71} &
  \textbf{178K} &
  0.59 &
  \textbf{0.11} &
  {\color[HTML]{000000} 0.70} &
  \textbf{3.86} &
  2.29 &
  1606 &
  1092 &
  90K &
  130K \\ \hline
\end{tabular}
\label{tab:hardware_performance}
\end{table*}

\subsection{Compare with Embedded GPUs}

\begin{figure}[t]
  \centering
  \includegraphics[width=\columnwidth]{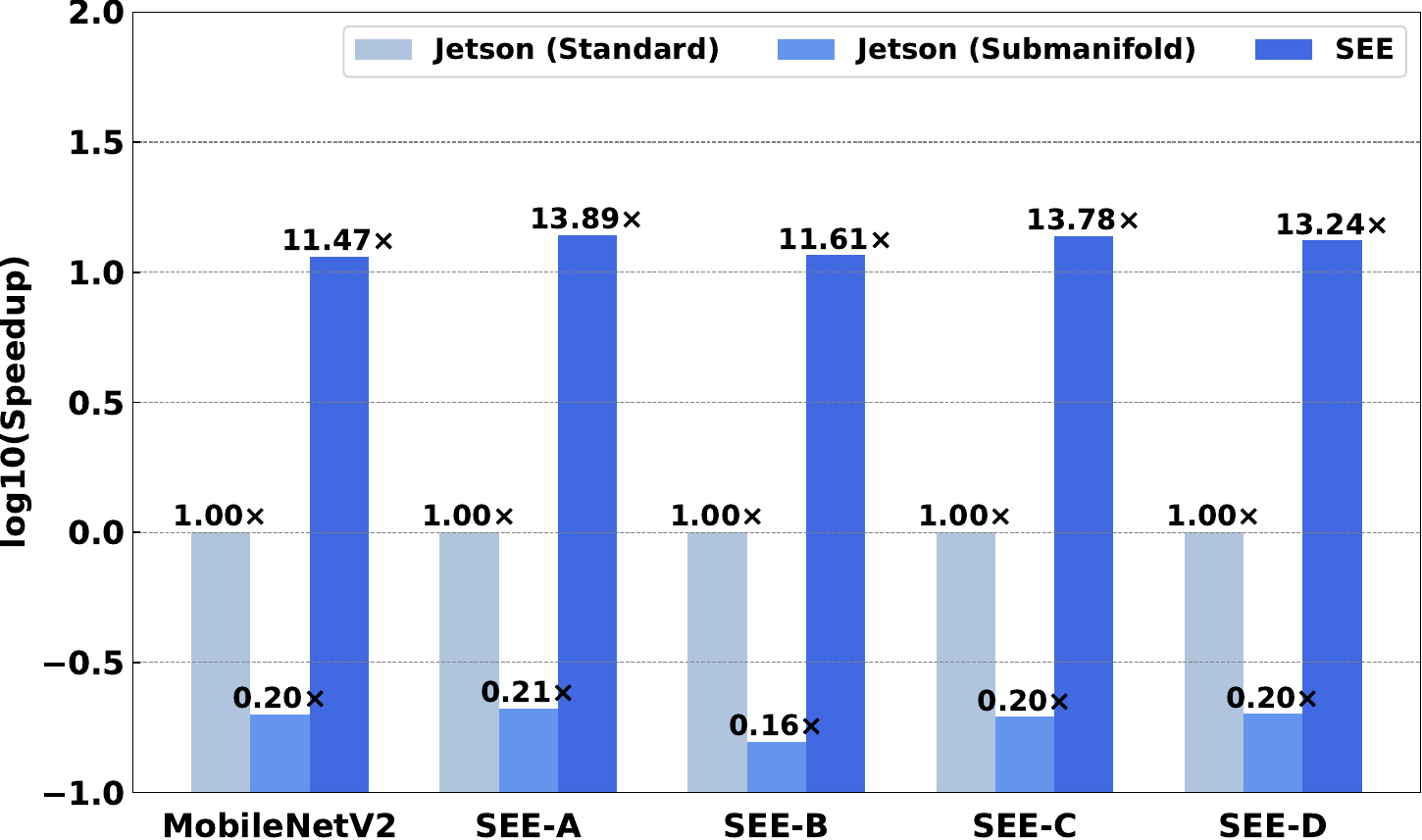}
  \caption{Latency speedup of \aname over an Nvidia Jetson GPU. We measure the latency (batch size = 1) for the standard and submanifold implementation using NVIDIA Jetson Xavier NX under MobileNetV2 and \aname-series network and calculate the speedup.}
  \label{fig:speedup}
\end{figure}

Finally, we conducted evaluations of our design using the NVIDIA Jetson Xavier NX, a widely-used embedded GPU. Similar to ESDA, we assessed both the dense DNN implementation using PyTorch and the submanifold sparse DNN implementation utilizing the MinkowskiEngine library. We calculated the average latency (batch=1) of the entire test set for the three settings, while the latency of standard implementation is defined as the baseline.

The results are shown in Figure \ref{fig:speedup}. Our \aname implementation achieves a notable speedup ranging from $11.47\times$ to $13.89\times$ compared with the standard one. While compared to the submanifold GPU implementation, the speedup can reach $57.4\times$, $66.1\times$, $72.6\times$, $68.9\times$, and $66.2\times$. The remarkable speedups highlight the significant efficiency improvement achieved by the co-designed hardware accelerator compared to both the standard and submanifold GPU implementations. The GPU implementation of submanifold sparse convolution typically exhibits slower performance than the standard dense baseline. This is primarily due to the significant overhead of sparse coordinate bookkeeping, particularly noticeable during batch 1 inference.



%% file: 10_conclusion.tex
\section{Conclusion and Future Work}
\label{sec:conclusion}

We present an efficient event-based eye-tracking solution called \aname through software/hardware co-design. \aname models utilize an SCNN backbond for feature extraction, followed by a GRU+FC component for temporal fusion and eye center localization. 
\aname system leverages the heterogeneous hardware resource of an embedded FPGA SoC platform and accelerates the SCNN using a novel sparse dataflow accelerator.
Furthermore, a hardware-software co-optimization framework is developed to obtain compact models optimal accuracy and latency tradeoffs. The results demonstrate impressive system performance, with a latency $0.6$ ms to $0.94$ ms for each prediction with around 99\% p10 accuracy. The overall latency speedups can reach $11.2\times$ to $72.6\times$ when compared to an embedded GPU.

Despite the outstanding performance \aname achieved, we aim to enhance the further latency performances by integrating the recurrent module or attention modules into our FPGA dataflow accelerator. This endeavor necessitates the development of novel quantization techniques or the implementation of some non-linear functions, as well as the support of inter-batch pipeline.

%% file: 12_appendix.tex
\section{Appendix Section}
\label{sec:appendix_section}
Supplementary material goes here.